\documentclass[conference,compsoc]{IEEEtran}
\ifCLASSOPTIONcompsoc
  \usepackage[nocompress]{cite}
\else
  \usepackage{cite}
\fi

\ifCLASSINFOpdf
\else
\fi

\usepackage{amsmath}
\usepackage{algorithmic}
\usepackage{array}
\usepackage{cite}
\usepackage[numbers,sort&compress]{natbib}
\usepackage{fixltx2e}
\usepackage{stfloats}
\usepackage{url}
\usepackage{tabularx}
\usepackage{graphicx}

\begin{document}

\title{SLIM LSTMs}


\author{\IEEEauthorblockN{Fathi M. Salem}
\IEEEauthorblockA{Circuits, Systems, and Neural Networks (CSANN) Lab\\Department of Electrical and Computer Engineering\\
Michigan State University\\
East Lansing, Micigan 48824--1226\\
Email:salemf@msu.edu}}

\maketitle

\begin{abstract}
Long Short-Term Memory (LSTM) Recurrent Neural networks (RNNs) rely on gating signals, each driven by a function of a weighted sum of at least 3 components: (i) one of an adaptive weight matrix multiplied by the incoming external input vector sequence, (ii) one adaptive weight matrix multiplied by the previous memory/state vector, and (iii) one adaptive bias vector. In effect, they augment the simple Recurrent Neural Networks (sRNNs) structure with the addition of a "memory cell" and the incorporation of at most 3 gating signals. 

The standard LSTM structure and components encompass redundancy and overly increased parameterization. In this paper, we systemically introduce variants of the LSTM RNNs, referred to as SLIM LSTMs. These variants express aggressively reduced parameterizations to achieve computational saving and/or speedup in (training) performance---while necessarily retaining (validation accuracy) performance comparable to the standard LSTM RNN.  
\end{abstract}

% \fi
\IEEEpeerreviewmaketitle

\section{Introduction}

There are now three main gating architectures for Recurrent Neural networks (RNNs) in “Deep Learning”, with impressive demonstrated performance in sequence-to-sequence applications \cite{DBLP:journals/corr/JohnsonSLKWCTVW16}, \cite{chung2014empirical}, \cite{gers2002learning}, \cite{hochreiter1997long}, \cite{zaremba2015empirical}, \cite{Odyssey2016} and \cite{bengio1994learning}. These are known as LSTM ([2-4]), GRU ([3]), and MGU ([4]) RNNs. LSTM RNN, the dominant workhorse in sequence processing, relies on three gating signals, GRU on two and MGU on one. Each gating signal is itself a replica of a simple recurrent neural network with its own parameters (at least \textit{two} matrices and a bias vector). Specifically, each gating signal is an output of a logistic nonlinearity driven by a weighted sum of at least \textit{three} terms: (i) one adaptive weight matrix multiplied by the incoming external vector sequence, (ii) one adaptive weight matrix multiplied by the previous memory/activation state vector, and (iii) one (adaptive) bias vector. This is the basic composition driving the gating mechanism in the gated RNN architectures literature. There are a host of variants starting from simple RNN (sRNN), to basic RNN (bRNN) \cite {salem2016basic}, to more complex gated variants, see, e.g., \cite{zaremba2015empirical} and \cite{Odyssey2016}.

\subsection{The rationale in developing the SLIM LSTMs:}

A key point is to (i) recognize and exploit the role of the internal dynamic "state" that captures the essential information about processing an input signal's time-history profile, and (ii), in time-series signal processing in recurrent systems, there is no need for repeated matrix multiplication of internal states beyond multiplying the (external) input sequence. As a matrix multiplication signifies scaling and rotation (say, mixing) of elements of a signal, one may use only scaling  of subsequent processing after the matrix multiplication of the input signal. Scaling can be expressed as a point-wise (Hadamard) multiplication. These two observations are exploited in defining the new family of SLIM architectures of the LSTMs.

As the state contains the essential summary information about a network, including the input sequence profile history, one can eliminate (redundant) terms not containing the state, directly or indirectly, in the gating signals. The gating signals' two weights and bias vector update laws depend on the external input signal and/or the previous memory state(s). Thus, there is redundancy in using all three terms to generate the gating signal(s) to achieve effective learning towards a desired (low loss or) high accuracy performance. Exploiting this observation allows for the development of several variant networks with reduced parameters resulting in computational savings. 

The view is to consider the gating as control "signals" which essentially only needs a measure of the network's state. In that view, the form of the standard LSTM network is overly redundant in generating such control signals. For example, it is redundant to provide the state (which may be represented by the memory cell or the activation unit, but not both!) and the external input signal to the gating  signal. 

For one, the derived (gradient-descent) update learning law(s) of the bias vector itself depends on the prior memory state vector and/or the (previous) external input vector. The state vector, again, captures all information pertaining to the signals in the dynamic system history profile--- specifically the external input prior (time-) sequence. A present input value may add a new (discounted)  information to prior state values; however, it may also bring an instantaneous outlier value corrupted by noisy measurements or external noise. On that basis, we assert forms that eliminate the instantaneous input sample from (all) gating signals. The intent is to strife to retain the accuracy performance of a gated RNN while aggressively reducing the number of (adaptive) parameters to various degrees. Such parameter reduced architectures would speedup execution in training and inference modes and may be more suitable for limited embedded or mobile computing platforms. 

From a recurrent dynamic systems view, the qualitative performance is expected to be retained. However, the quantitative performance would of course vary to various degrees as the number of parameters is reduced to various different levels. 
\\

This paper collectively presents the network families for SLIM LSTM RNNs and shows the interconnections among them. Let us denote the dimension of the input vector (sequence) to be $m$, and dimension of the hidden unit, and similarly the state (or memory) vector to be $n$.
We now introduce and overview some of the new gating variants as follows: 
\\
\\
\textbf{Variant1:} from all gating signals, remove the external signals and associated weight matrix. This amounts to reducing the parameters, per gate, by $n \times m$. Alternatively, the existing parameters, per gate, are $n^2 +n$.
\\
\\
\textbf{Variant2:} from all gating signals, remove the external signals and their associated weight matrix, and remove the bias vector. This amounts to reducing the parameters, per gate, by  
$n \times m+n = n(m+1)$.  Alternatively, the existing parameters, per gate, are $n^2$.
\\
\\
\textbf{Variant3:} from all gating signals, remove the external signals and their associated weight matrix, and remove the previous memory/state and their associated weight matrix. This leaves only the bias vector. This amounts to reducing the parameters, per gate, by  $n \times m+n \times n=n(n+m)$. Alternatively, the existing parameters, per gate, are $n$.
\\

The removal of any such parameters eliminates the adaptive computational effort for estimating them, and the need to store them or any intermediate steps in the adaptive process.  To appreciate this reduction, the breakthrough application in language translation [1] uses 4 to 8 cascaded LSTM RNNs. This translates to the requirement of less memory and less CPU/GPU resources which would lead to faster training and learning, and potentially allow for even more scaled systems. 
\\
\\
\textbf{Variant4:} Same as Variant 2; however, matrix multiplication is replaced by point-wise (Hadamrd) mulitplication for the previous hidden state vector. This further reduces the parameters, per gate, to 
$n \times m+n+(n^2-n) = n \times m +n^2$.  Alternatively, the existing parameters, per gate, are $n$.
\\
\\
\textbf{Variant5:} Same as Variant 1; however, matrix multiplication is replaced by point-wise (Hadamrd) mulitplication for the previous hidden state vector. This reduces the parameters, per gate, to 
$n \times m+(n^2-n) = n \times m +n^2-n$.  Alternatively, the existing parameters, per gate, are $2n$.
\\

More variants will be described in the following sections.  We will describe a diverse set of variant networks with the intended goal of providing a host of choices, balancing parameter-reduction and quantitative performance in (validation-testing) accuracy. We have already demonstrated the quantitative performances of these new network variants in recent publications (\cite {LuSalem2017, HeckSalem2017, DeySalem2017, AkandehSalem2017I, AkandehSalem2017II,  AkandehSalem2017III})--- albeit for initial datasets. Here, we describe the insight and reasoning into the reduced networks' developments in a comprehesive way \cite{salem2016reduced}. We indicate how those network variants link the simple RNN in graded complexity all the way to the full \textit{standard} LSTM network. 

\section{Background: The Simple and LSTM RNNs}
The so-called simple RNN has a recurrent hidden state as in
\begin{align} \label{eqn:1}
	h_t = g(Wx_t + Uh_{t-1} + b) 
\end{align}
where $x_t$ is the (external)  $m-$dimensional input vector at time $t$, $h_t$ the $n-$dimensional hidden state, $g$ is the (point-wise) activation function, such as the logistic function,  the hyperbolic tangent function, or the rectified Linear Unit (ReLU) \cite{chung2014empirical, zaremba2015empirical}, and $W,\;U\; and\; b$ are the appropriately sized parameters (namely, two weights and a bias). Specifically, in this case,  $W$ is an $n \times m$ matrix,  $U$ is an $n \times n$ matrix, and $b$ is an $n \times 1$ matrix (or vector).

Bengio et al. \cite{bengio1994learning} showed that it is difficult to capture long-term dependencies using such simple RNN because the (stochastic) gradients tend to either vanish or explode with long sequences.  The Long Short-Term Memory (LSTM) RNN \cite{hochreiter1997long, gers2002learning} has been the first network proposed to mitigate the “vanishing” or “exploding” gradient problems. 

\subsection{The Long Short-Term Memory (LSTM) RNN}
The LSTM RNN architecture introduces the "memory cell" to augment the simple RNN architecture of equation (\ref{eqn:1}). Further, it introduces the gating (control)  signals to basically incorporate the previous memory value to the new computaions. Let the simple RNN computation produce its contribution to an intermediate variable, say $\tilde{c}_t$,  and add it in a  weighted-sum (element-wise) to the previous value of the internal memory state, say $c_{t-1}$, to produce the current value of the memory cell (state) $c_t$.  These operations are expressed as the following set of discrete dynamic equations:
\begin{align}
	\label{eqn:2}	\tilde{c}_t &= g(W_c x_t + U_c h_{t-1} + b_c)\\
	\label{eqn:3}	 c_t &= f_t \odot c_{t-1} + i_t \odot \tilde{c}_t\\
	\label{eqn:4}	h_t &= o_t \odot g(c_t)
\end{align}
The weighted sum is implemented in Eqn (\ref{eqn:3}) as element-wise (Hadamard) multiplication denoted by $\odot$ to gating (control) signals $ i_t$ and $ f_t$, respectvely . The gating signals $i_t,\; f_t \; and\; o_t$ denote, respectively, the \textit{input}, \textit{forget}, and \textit{output} gating signals at (discrete) time or step $t$ \cite{hochreiter1997long, Odyssey2016}.  In Eqns (\ref{eqn:2}) and (\ref{eqn:4}), the activation nonlinearity $g$ is typically the hyperbolic tangent function, however other forms are possible, e.g., the logistic function or the rectified Linear Unit (reLU).

These control gating signals are in fact replica of the basic equation (\ref{eqn:1}), with their own replica parameters and simply replacing $g$  by the logistic function. The logistic function limits the gating signals to within $0$ and $1$. The specific mathematical form of the gating signals are thus expressed as the vector equations:
\begin{align}
	i_t &= \sigma(W_i x_t + U_i h_{t-1} + b_i)\\
	f_t &= \sigma(W_f x_t + U_f h_{t-1} + b_f)\\
	o_t &= \sigma(W_o x_t + U_o h_{t-1} + b_o)
\end{align}
where $\sigma$ is the logistic nonlinearity and the parameters for each gate consist of two matrices and a bias vector. Thus, the total number of parameters (represented as matrices and bias vectors) for the 3 gates and the memory cell structure are, respectively, $W_i,\; U_i,\; b_i,\; W_f,\; U_f,\; b_f,\; W_o,\; U_o,\; b_o,\; W_c,\; U_c$ and $b_c$. These parameters are all updated at each training step (or mini-batch) and stored. It is immediately noted that the number of parameters in the LSTM model is increased 4-folds from the simple RNN model in Eqn (\ref{eqn:1}). Assume that the cell state $c_t$ is $n$-dimensional. (Note that the activation and all the gates have the same dimensions). Assume also that the input signal is $m$-dimensional. Then, the total parameters in the LSTM RNN is equal to $4 \times (n^2 + nm +n)$.

\section{SLIM LSTMs: reduction within gates}

The gating signals in Gated RNNs enlist all of (i) the previous hidden unit or state, (ii) the present input signal, and (iii) a bias, in order to enable the Gated RNN to essentially learn sequence-2-sequence mappings. The dominant adaptive algorithms used in training are varieties of backpropagation through time (BPTT)  stochastic gradient descen.  The gates, each, simply replicates a simple RNN. All parameters in this LSTM sturuture are updated using the BPTT stochastic gradient descent to minimize a loss function \cite{gers2002learning, Odyssey2016}. The concept of state, which in essence summarizes the information of the Gated RNN up to the present (or previous) time step, contains the information about the profile of the input sequence. Moreover, the parameter update also includes information pertaining to the state (and co-state) of the overall network structure  \cite{salem2016reduced,salem2016basic}. 

For tractable and modular realizations, we consider applying the modifications to all gating signals uniformaly. Thus we can consider only the modifications to one of the gating signals, say the i-th gating signal, and replicate the modifications in all other gating signals.

A gating signal is driven by 3 components, resulting in 8 possible variations--- including the trivial one when all three components are absent. without the external input signal, there 3  non-trivial variants per gate. For efficiency, we consider the 3 variants without the external input sequence as the input sequence over its time/sample horizon is contained in the "state." 
\\
\subsection{Variant 1: The LSTM\_1 RNN}
In this variant, each signal gate is computed using the previous hidden state and the bias, thus reducing the total number of parameters from the 3 gate signals, in comparison to the LSTM RNN, by $3 \times nm$. 
\begin{align}
	i_t &= \sigma(U_i h_{t-1} + b_i)\\
	f_t &= \sigma(U_f h_{t-1} + b_f)\\
	o_t &= \sigma(U_o h_{t-1} + b_o)
\end{align}

\subsection{Variant 2: The LSTM\_2 RNN}
In this variant, each signal gate is computed using only the previous hidden state, thus reducing the total number of parameters from the 3 gate signals, in comparison to the LSTM RNN, by $3 \times (nm + n)$.
\begin{align}
	i_t &= \sigma(U_i h_{t-1})\\
	f_t &= \sigma(U_f h_{t-1})\\
	o_t &= \sigma(U_o h_{t-1})
\end{align}

\subsection{Variant 3: The LSTM\_3 RNN}
In this variant, each gate is computed using only the bias, thus reducing the total number of parameters in the 3 gate signals, in comparison to the LSTM RNN, by $3 \times (nm + n^2)$.

\begin{align}
	i_t &= \sigma(b_i)\\
	f_t &= \sigma(b_f)\\
	o_t &= \sigma(b_o)
\end{align}

In order to reduce the parameters even further, one replaces the standard multiplications by point-wise multiplications. In the case of the hidden units, the matrices $U_{*}$ are reduced into (column) vectors of the same dimension as the hidden units (i.e., $n$). We denote these corresponding vectors by $u_{*}$ as delineated next.  

\subsection{Variant 4: The LSTM\_4 RNN}
In this variant, each gate is computed using only the previous hidden state but with point-wise multiplication. Thus one reduces  the total number of parameters, in comparison to the LSTM RNN, by $3 \times (nm + n^2)$.

\begin{align}
	i_t &= \sigma(u_i \odot  h_{t-1})\\
	f_t &= \sigma(u_f \odot h_{t-1})\\
	o_t &= \sigma(u_o \odot  h_{t-1})
\end{align}

\subsubsection{Variant 4i: The LSTM\_4i RNN}
In this variant, only the (so-called) input (or update) gate is computed, thus further reducing the total number of parameters.

\begin{align}
	i_t &= \sigma(u_i \odot  h_{t-1} )\\
	f_t &= \alpha,  ~~~~~ 0 \leq  | \alpha | \leq  1 \\
	o_t &= 1
\end{align}
$\alpha$ is typically a constant between between $0.5$ and $0.99$ in order to stabilize the (gated) RNN--- in a Bounded Input Bounded Output (BIBO) sense \cite{salem2016basic}.
This model reduces to the more compact form: 
\begin{align}
	\label{eqn:4431}	 c_t &=  \alpha ~ c_{t-1} + i_t \odot g(W_c x_t + U_c h_{t-1} + b_c)\\
	\label{eqn:4441}	h_t &=  g(c_t)
\end{align}
where $c_t$ is clearly the only state of the network, and the activation, $h_t$ is a (nonlinear) function of the state. This is in contrast to some claims in the literature that consider both $c_t$ and $h_t$, togther, as states of the network!

\subsubsection{Variant 4ib: The LSTM\_4ib RNN}
Motivated by the bRNN model in \cite {salem2016basic}, we can remove the nonlinearity in eqn [\ref{eqn:4431}], and thus use the "equivalent" dynamic architecture with a single activation function {$g(.)$, namely, 
\begin{align}
	\label{eqn:431}	c_t &=  \alpha ~ c_{t-1} + i_t \odot (W_c x_t + U_c h_{t-1} + b_c)\\
	\label{eqn:441}	h_t &=  g(c_t)
\end{align}
\\
\subsection{Variant 5: The LSTM\_5 RNN}
In this variant, each gate is computed using only bias plus the previous hidden state with point-wise multiplication as follows.

\begin{align}
	i_t &= \sigma(u_i \odot  h_{t-1} + b_i)\\
	f_t &= \sigma(u_f \odot h_{t-1} + b_f)\\
	o_t &= \sigma(u_o \odot  h_{t-1} + b_o)
\end{align}

Analogous to the previous subsection, we reduce the gating signals further. 

\subsubsection{Variant 5i: The LSTM\_5i RNN}
In this variant, only the input gate is used. The other gates are set to constants as follows:
\begin{align}
	i_t &= \sigma(u_i \odot  h_{t-1} + b_i)\\
	f_t &= \alpha , ~~~~~  0 \leq |\alpha| \leq  1 \\
	o_t &= 1
\end{align}
$\alpha$ has absolute value less than or equal to 1 for bounded input bounded output (BIBO) stability, but typically is set as a (hyperparameter) constant between $0.5$ and $0.99$. 
This model reduces to the more compact form: 
\begin{align}
	\label{eqn:5531}	c_t &=  \alpha ~ c_{t-1} + i_t \odot g(W_c x_t + U_c h_{t-1} + b_c)\\
	\label{eqn:5541}	h_t &=  g(c_t)
\end{align}

\subsubsection{Variant 5ib: The LSTM\_5ib RNN}
Again, motivated by the basic RNN (bRNN) model in \cite {salem2016basic}, we can remove the nonlinearity in eqn [\ref{eqn:5531}], and thus use the "equivalent" dynamic architecture with a single activation function {$g(.)$, namely, 
\begin{align}
	\label{eqn:531}	c_t &=  \alpha ~ c_{t-1} + i_t \odot (W_c x_t + U_c h_{t-1} + b_c)\\
	\label{eqn:541}	h_t &=  g(c_t)
\end{align}
\\

\subsection{Variant 6: The LSTM\_6 RNN}
In this variant, each gate is computed using only constants.
\begin{align}
	i_t &= 1 \\
	f_t &= \alpha , ~~~~~  0 \leq |\alpha| \leq  1 \\
	o_t &= 1
\end{align}

In {Variant\_ 6}, the overall system equations can compactly be expressed as

\begin{align}
	\label{eqn:611}   c_t &= \alpha ~ c_{t-1} + g(W_c x_t + U_c h_{t-1} + b_c) \\
	\label{eqn:612}	 h_t &=  g(c_t)
\end{align}

\subsubsection{Variant 6b: The LSTM\_6b RNN}
Again, motivated by the basic RNN (bRNN) network in \cite {salem2016basic}, we can remove the nonlinearity in eqn [\ref{eqn:611}], and thus use the "equivalent" dynamic architecture with a single activation function {$g(.)$, namely, 

\begin{align}
	\label{eqn:611}   c_t &= \alpha ~ c_{t-1} + (W_c x_t + U_c h_{t-1} + b_c) \\
	\label{eqn:612}	 h_t &=  g(c_t)
\end{align}

This network is in effect the bRNN model reported in \cite {salem2016basic} with the input vector advanced by one sample. 
\\
\\
\section{SLIM LSTMs: reduction in gates and the memory cell input block}

We next apply the reduction to the body of the simple RNN (sRNN) network as the input block within the standard LSTM equations, namely:
\begin{align}
	\label{eqn:4442}	\tilde{c}_t &= g(W_c x_t + U_c h_{t-1} + b_c)
\end{align}
\\
It is observed that the external input signal has its entry point to the the LSTM for processing. Its ``mixing" matrix, i.e., $W_c $, is needed for full transformation (scaling and rotation) of the external signal $x_t$, the bias parameter $b_c$ would likley be needed in case the external signal does not have zero mean. However, the $n \times n$-matrix $U_c$ may be replaced by an $n-d$-vector to retain scaling (point-wise) but not rotation. The main observation is that in propagation over the time horizon, each instant of the  vector $\tilde{c}_t$ will be a function of a weighted sum of all components of the external input signal. Thus all ``state-vector'' components will be ``mixed'' naturally due the mixing of the external input signal. Thus, one can reduce the parameterization from 
$n^2$ to $n$, and consequently reducing all associated update computation and storage for $n^2-n$ parameters.  For this one matrix, the reduction is $100(1-1/n)\%$. For n-d LSTM, this becomes $99 \%$ reduction!
\\
\\ 
The new variants are focusing on the ``memory cell input block'' of Eqn (\ref{eqn:4442}).  One leaves the multiplication in the first term that contains the input sequence unchanged, in order to provide mixing multiplication to the incoming input sequence.  Here, one only alters the term involving the activation unit $h_{t-1}$ into the $g$ function. The multiplication here can be  made point-wise (Hadamard) multiplication which provides scaling but no rotation. The rationale is that the "state" (namely, the memory cell  $c_t$, and consequently the activation $h_t$), over the sequence horizon, integrates mixtures of the components of the input sequence, and therefore, there is apparent redundancy in further rotations the states. Thus, it is a candidate for point-wise (scaling only) Hadamard multiplication in order to reduce parameterization (while preserving potential performance). \\
\\
\\
The actions here can generate additional possibilities when counting the possiblities of the presence and absence of each term in comparison to the baseline original LSTM form. The ``memory cell input block'' equation can generate a total of $2^2=4$ variants including the baseline, or 3 new variants. We choose two Cell variants below as follows:

\subsection{Variant Cell 1}
Here , one replaces the original $n \times n$-matrix $U_c$ by the $n-d$-vector $u_c$, and applies point-wise (Hadamard) multiplication to the previous hidden activation $h_{t-1}$. The bias parameter is also present; note that, in the paper, the bias parameter is present in odd-numbered variants. 
\begin{align}
	\label{eqn:2000}	\tilde{c}_t &= g(W_c x_t + u_c \odot h_{t-1} + b_c)\\
	\label{eqn:3000}	 c_t &= f_t \odot c_{t-1} + i_t \odot \tilde{c}_t\\
	\label{eqn:4000}	h_t &= o_t \odot g(c_t)
\end{align}

\subsection{Variant Cell 2}
Here, one replaces the original $n \times n$-matrix $U_c$ by the $n-d$-vector $u_c$, and applies point-wise (Hadamard) multiplication to the previous hidden activation $h_{t-1}$. The bias parameter is removed; note that, in his paper, the bias parameter is removed in even-numbered variants. 
\begin{align}
	\label{eqn:2000}	\tilde{c}_t &= g(W_c x_t + u_c \odot h_{t-1})\\
	\label{eqn:3000}	 c_t &= f_t \odot c_{t-1} + i_t \odot \tilde{c}_t\\
	\label{eqn:4000}	h_t &= o_t \odot g(c_t)
\end{align}

 It is noted that one may consider these variants in combination (or linked) with the variations introduced on the gating signals to obtain the total possible diverse variations. As an example, we introduce the following reduced variations involving a combination of gating signals and ``memory cell" input block reduced paramerization. In the listed variations below, we retain the same variation numbering as before preceeded by the letter $C$ to signify that these variants are alterations including the ``memory cell''  input block.

\subsection{Variant C3: The LSTM\_C3 RNN}
In this variant, each signal gate is computed using the previous hidden state and the bias, thus reducing the total number of parameters from the 3 gate signals, in comparison to the LSTM RNN, by $3 \times nm$. 
\begin{align}
	i_t &= \sigma(b_i)\\
	f_t &= \sigma(b_f)\\
	o_t &= \sigma(b_o)
\end{align}
with the ``memory cell'' reduced form
\begin{align}
	\label{eqn:C202}	\tilde{c}_t &= g(W_c x_t + u_c \odot  h_{t-1} + b_c)\\
	\label{eqn:C302}	 c_t &= f_t \odot c_{t-1} + i_t \odot \tilde{c}_t\\
	\label{eqn:C402}	h_t &= o_t \odot g(c_t)
\end{align}

\subsection{Variant C4: The LSTM\_C4 RNN}
In this variant, each signal gate is computed using the previous hidden state and the bias, thus reducing the total number of parameters from the 3 gate signals, in comparison to the LSTM RNN, by $3 \times nm$. 

\begin{align}
	i_t &= \sigma(u_i \odot  h_{t-1})\\
	f_t &= \sigma(u_f \odot h_{t-1})\\
	o_t &= \sigma(u_o \odot  h_{t-1})
\end{align}
with the ``memory cell'' reduced form
\begin{align}
	\label{eqn:CC202}	\tilde{c}_t &= g(W_c x_t + u_c \odot  h_{t-1} + b_c)\\
	\label{eqn:CC302}	 c_t &= f_t \odot c_{t-1} + i_t \odot \tilde{c}_t\\
	\label{eqn:CC402}	h_t &= o_t \odot g(c_t)
\end{align}

\subsubsection {Variant C4i: The LSTM\_C4i RNN}

In this variant, only the (so-called) input (or update) gate is computed, thus reducing the total number of parameters.

\begin{align}
	i_t &= \sigma(u_i \odot  h_{t-1} )\\
	f_t &= \alpha,  ~~~~~ 0 \leq  | \alpha | \leq  1 \\
	o_t &= 1
\end{align}
$\alpha$ is typically a constant between between $0.5$ and $0.99$ to stabilize the (gated) RNN. 
This model reduces to the more compact form: 
\begin{align}
	\label{eqn:CC31}	 c_t &=  \alpha ~ c_{t-1} + i_t \odot g(W_c x_t + u_c \odot  h_{t-1} + b_c)\\
	\label{eqn:CC41}	h_t &=  g(c_t)
\end{align}

\subsubsection{Variant C4ib: The LSTM\_C4ib RNN}

Motivated by the bRNN model in \cite {salem2016basic}, we can remove the nonlinearity in eqn [\ref{eqn:4431}], and thus use the "equivalent" dynamic architecture with a single activation function {$g(.)$), namely, 
\begin{align}
	\label{eqn:CC431}	c_t &=  \alpha ~ c_{t-1} + i_t \odot (W_c x_t + u_c \odot  h_{t-1} + b_c)\\
	\label{eqn:CC441}	h_t &=  g(c_t)
\end{align}
\\

\subsection{Variant C5: The LSTM\_C5 RNN}
In this variant, each gate is computed using only bias plus the previous hidden state with point-wise multiplication as follows.

\begin{align}
	i_t &= \sigma(u_i \odot  h_{t-1} + b_i)\\
	f_t &= \sigma(u_f \odot h_{t-1} + b_f)\\
	o_t &= \sigma(u_o \odot  h_{t-1} + b_o)
\end{align}
with the ``memory cell'' reduced form
\begin{align}
	\label{eqn:C5C202}	\tilde{c}_t &= g(W_c x_t + u_c \odot  h_{t-1} + b_c)\\
	\label{eqn:C5C302}	 c_t &= f_t \odot c_{t-1} + i_t \odot \tilde{c}_t\\
	\label{eqn:C5C402}	h_t &= o_t \odot g(c_t)
\end{align}

Now, we reduce the gating signals. 

\subsubsection{Variant C5i: The LSTM\_C5i RNN}

In this variant, only the input gate is used. The other gates at set to constants. each gate is computed using only the bias.
\begin{align}
	i_t &= \sigma(u_i \odot  h_{t-1} + b_i)\\
	f_t &= \alpha , ~~~~~  0 \leq |\alpha| \leq  1 \\
	o_t &= 1
\end{align}
$\alpha, ~ |\alpha | <= 1$,which is typically a constant between $0.5$ and $0.96$ to stablize the (gated) RNN. 
This model reduces to the more compact form: 56 
\begin{align}
	\label{eqn:5iC31}	 c_t &=  \alpha ~ c_{t-1} + i_t \odot g(W_c x_t + u_c \odot  h_{t-1} + b_c)\\
	\label{eqn:5iC41}	h_t &=  g(c_t)
\end{align}

\subsubsection{Variant C5ib: The LSTM\_C5ib RNN}

Again, motivated by the basic RNN (bRNN) model in \cite {salem2016basic}, we can remove the nonlinearity in eqn [\ref{eqn:5531}], and thus use the "equivalent" dynamic architecture with a single activation function {$g(.)$), namely, 
\begin{align}
	\label{eqn:5ibC531}	c_t &=  \alpha ~ c_{t-1} + i_t \odot (W_c x_t + u_c \odot  h_{t-1} + b_c)\\
	\label{eqn:5ibC541}	h_t &=  g(c_t)
\end{align}
\\

\subsection{Variant C6: The LSTM\_C6 RNN}
In this variant, each gate is computed using only constants.
\begin{align}
	i_t &= 1 \\
	f_t &= \alpha , ~~~~~  0 \leq |\alpha| \leq  1 \\
	o_t &= 1
\end{align}

In Variant C6, the overall system equations can compactly be expressed as

\begin{align}
	\label{eqn:11}   c_t &= \alpha ~ c_{t-1} + g(W_c x_t + u_c \odot h_{t-1} + b_c) \\
	\label{eqn:12}	 h_t &=  g(c_t)
\end{align}

\subsubsection{Variant C6b: The LSTM\_C5ib RNN}

Again, motivated by the basic RNN (bRNN) model in \cite {salem2016basic}, we can remove the nonlinearity in eqn [\ref{eqn:5531}], and thus use the "equivalent" dynamic architecture with a single activation function {$g(.)$), namely, 
\begin{align}
	\label{eqn:111}   c_t &= \alpha ~ c_{t-1} + (W_c x_t + u_c \odot h_{t-1} + b_c) \\
	\label{eqn:112}	 h_t &=  g(c_t)
\end{align}
As before, $\alpha$ is a (hyper-parameter) constant typically between between $0.5$ and $0.96$ for BIBO stability.
\section*{Remarks:}

1) Particularly in this last variant, as the parameters have been aggressively reduced, the (state) dimension of the network is a critical hyper-parameter that can be increased in order to increase the capacity of performance of the overall variant network. 
\\

2) in this variant, as in all other variants, the weight matrix $W_c$ may be replaced with a convolution kernel operations or suitable size.

\section{Concluding Remarks}
We have introduced a mosaic of new recurrent architectures that \textit{slim} down the standard LSTM architecture to simpler forms. We have eliminated the redundancy of parameters and in some cases allow for the presence of a single nonlinearity in the memory-cell recurrent network structure. As recurrent neural networks are employed to learn sequence-to-sequence mappings, the question turns to \textit{capacity}, i.e., the existence of a set of parameters in a given architecture (or variant) that enables approximate mappings of finite sequence-to-sequence using the training data--- while generalizing on validation/test datasets. In that context, the dimension of the variant would become an important (hyper-) parameter for some SLIM LSTMs that could be used to increase the capacity of improved performance. 

Many of these variants have already been validated to produce comparable performance to the standard LSTM RNN in recent publications \cite {LuSalem2017, HeckSalem2017, DeySalem2017, AkandehSalem2017I, AkandehSalem2017II,  AkandehSalem2017III, salem2016reduced}.
The remaining ones are currently being investigated in case studies. 

\section*{Acknowledgement}

This work was supported in part by the National Science Foundation under grant No. ECCS-1549517.

% Bibliography
\bibliographystyle{IEEEtranN}
\bibliography{SLIM_LSTMs_aut_Fathi_Salem}

\end{document}